\documentclass{article}
\usepackage{spconf,amsmath,graphicx}
\usepackage{amsfonts}
\usepackage[ruled,vlined]{algorithm2e}
\usepackage{booktabs}
\usepackage[flushleft]{threeparttable}
\SetKwComment{Comment}{$\triangleright$\ }{}

\usepackage{bbm}
\usepackage[hidelinks]{hyperref}

\title{Group Personalized Federated Learning}

\name{Zhe Liu, Yue Hui, Fuchun Peng}
\address{Meta AI, Menlo Park, CA, USA}

\begin{document}
\ninept
\maketitle

\begin{abstract}
Federated learning (FL) can help promote data privacy by training a shared model in a de-centralized manner on the physical devices of clients. In the presence of heterogeneous distributions of local data, personalized FL strategy is introduced to mitigate the potential client drift. In this paper, we present the group personalization approach for applications of FL in which there exist inherent partitions over clients that are significantly distinct. In our approach, the global FL model is fine-tuned through another FL training process over each homogeneous group of clients, after which each group-specific FL model is further adapted and personalized per client. The proposed method can be well interpreted from a Bayesian hierarchical modeling perspective. With experiments on two real-world datasets, we demonstrate this approach can achieve superior personalization performance than other FL counterparts.
\end{abstract}

\begin{keywords}
Federated learning, personalization, language modeling
\end{keywords}

\section{Introduction}
In recent years, there has been a rise in the popularity of a distributed learning technique called federated learning (FL) \cite{konevcny2016federated, konevcny2016federated2, mcmahan2017communication}. FL has been applied in many fields including recommendation \cite{chen2018federated}, smart keyboard suggestion \cite{arnold2016suggesting, ji2019learning}, keyword spotting \cite{leroy2019federated}, health care \cite{xu2019federated}, and automatic speech recognition (ASR) \cite{dimitriadis2020federated, guliani2021training, cui2021federated}.

FL can help promote data privacy by training a shared model in a de-centralized manner on users' local devices, so that raw data stays on physical devices. Specifically, FL distributes the training process among a large number of client devices, with each client device learning from private data and calculating model updates independently, then uploading those updates to a central server for aggregation. The updated model will later be delivered to each client device, after which this procedure is repeated until convergence.

The vanilla FL approach faces challenges in the presence of highly heterogeneous local data distributions. The personalized FL strategy seeks to address such performance issue and mitigate the potential client drift \cite{fallah2020personalized, li2021ditto, chen2022active}. Particularly, a two-step ``global FL training + local fine-tuning'' method is commonly adopted for personalization, where the trained global FL model is personalized for each FL client. This is done through a local adaptation step that involves additional training on each local dataset \cite{mansour2020three, tan2022towards}.

However, this two-step federated personalization approach has limitations when a majority of users only have a few training examples, which is common in practice due to long-tailed skewed distributions of user data. Fine-tuning a large global FL model on insufficient personal data may not improve the performance for individual clients or tends to suffer from overfitting.

For applications where there exist inherently partitioned groups among clients, each client can leverage the extra knowledge, learned from the training records of other clients in their group, and enhance their own personalized model. This procedure should also be conducted in a FL framework since raw data has to stay on devices.

In this paper, we present a novel three-step ``global FL training + group FL fine-tuning + local personalization'' approach. Specifically, it firstly follows the general FL training process where a single global FL model is learned. Then this trained global model is fine-tuned through another FL training process over each homogeneous group of clients. Finally, each group-level model is further adapted and personalized using the private data per client.

Our work mainly makes the following technical contributions: (1) proposing group personalized FL, an effective approach for integrating global aggregation, group-level knowledge sharing, and local training; (2) interpreting the proposed procedure from a Bayesian hierarchical modeling perspective; and (3) evaluating on real-world datasets for language modeling task, which achieves improved personalization results.

The rest of the paper is organized as follows. We review related work in Section~\ref{review}. Section \ref{methodology} presents the proposed method of group personalized FL. Section \ref{properties} interprets the presented procedure from a Bayesian hierarchical modeling perspective. Section \ref{experiments} shows the experiments on two real-world datasets. We conclude in Section \ref{conclusion}.

\section{Related Work}
\label{review}
Recently, there is an emerging line of research that develops clustering strategies for clients in the FL settings \cite{ghosh2020efficient, sattler2020clustered, briggs2020federated, xie2020multi, wainakh2020enhancing, duan2021fedgroup, biswas2021privacy}. Particularly, previous work in \cite{wainakh2020enhancing} and \cite{biswas2021privacy} proposes to apply FL on a hierarchical architecture and explores the potential benefits of using it to address privacy-related issues. Authors in \cite{ghosh2020efficient} present an iterative clustering algorithm which estimates the cluster identities of the clients and optimizes model parameters for the clusters. Another approach \cite{sattler2020clustered,duan2021fedgroup} groups the training of clients based on the similarities between the clients' optimization directions. Moreover, paper of \cite{xie2020multi} introduces a multi-center aggregation mechanism which learns multiple global models from data, and simultaneously derives the optimal matching between clients and centers.

As most of existing literature focuses on clustering algorithms over clients, our work mainly investigate how the group or cluster information can be efficiently utilized for improving personalization performance. To the best of our knowledge, our work is the first that provides an empirical study on combining group or cluster based FL with personalization. While our paper mainly investigates the use of group information for enhancing personalized FL, the comparison of various clustering algorithms for inferring the groups of clients is beyond the scope of this work.

\section{Group Personalized FL}
\label{methodology}
In this section, we present group personalized FL, which is a three-step method consisting of global FL training, group FL fine-tuning, and local personalization. %The entire procedure is outlined in Algorithms~\ref{algo1}, \ref{algo2}, and \ref{algo3}, with more details in subsections \ref{method:1}, \ref{method:2}, and \ref{method:3}.

\subsection{Global FL Training}
\label{method:1}
The proposed group personalized FL starts with a general FL training until convergence. Suppose at round $t$ of FL training, each selected client downloads the model $\Theta_t$ from server and performs secure local training on their own device. Mini-batch stochastic gradient descent (\texttt{SGD}) can be used as the local optimizer with learning rate $\eta_l$. After $K$ epochs of training, the client uploads its model update $\Delta_i^t$ (i.e. difference of model parameters) to the central server over a secure connection.

\begin{algorithm}
\SetAlgoLined
 Hyper-parameters $T, K, \eta_l, \eta_G$\;
 Initialize $\Theta_1$\;
 \For{\emph{each round} $t=1,2,\ldots, T$}{
  Deliver $\Theta_t$ to each client\\
  Sample a subset $\mathcal{I}_t$ of clients\\
  \For{\emph{each client} $i\in\mathcal{I}_t$ \emph{in parallel}}{
    Load $\theta_{i,1}^t:=\Theta_t$\\
    Train $K$ epochs via \texttt{SGD}($\theta_{i,k}^t$, $\eta_l$)\\
    Send $\Delta_i^t:=\Theta_t-\theta_{i,K+1}^t$ to server\\
  }
  $\Theta_{t+1}\leftarrow\texttt{FedAdam}(\Theta_t, \Delta_{\mathcal{I}_t}, \eta_G)$\\
 }
 Emit $\Theta_{T+1}$;
 \caption{Global FL Training.}
 \label{algo1}
\end{algorithm}

Then once the central server receives all model updates $\Delta_{\mathcal{I}_t}:=\{\Delta_i^t\}_{i\in {\mathcal{I}_t}}$ from selected clients of $\mathcal{I}_t$, it computes the averaged model difference or ``pseudo-gradient'' which will be used in server optimizer update. The \texttt{FedAdam} optimizer \cite{reddi2020adaptive} can be used for updating the global model with $\eta_G$ being the learning rate. Algorithm~\ref{algo1} depicts the client-side and server-side updates in global FL training, which lead to the single model of $\Theta_{T+1}$ upon convergence.

\subsection{Group FL Fine-Tuning}
\label{method:2}
In the scenarios where the FL clients can be partitioned into different groups, such cluster information can be utilized to help mitigate the heterogeneity across different groups and also enhance personalization performance through within-group knowledge sharing among clients for each group. These groups might exist naturally or can be inferred from data. For example, authors in \cite{sattler2020clustered} uses cosine similarity of the gradient updates of the clients to partition clients into groups. Please see Section~\ref{review} for a discussion of clustering methods.

For any group $g$ of clients, the trained global FL model $\Theta_{T+1}$ is further fine-tuned in the FL framework, namely group FL fine-tuning. Specifically, at round $t$ of group FL fine-tuning, each selected client in group $g$ downloads model $\Theta_{g,t}$ from server and trains on their private data for $K$ epochs. Once the central server receives all model updates from selected clients within set $\mathcal{I}_{g,t}$ in group $g$, group-specific model update is executed and thus leads to $\Theta_{g,t+1}$. After training for $T_g$ rounds, $\Theta_{g,T_g+1}$ is obtained for group $g$. This process is summarized in Algorithm~\ref{algo2}.

\begin{algorithm}
\SetAlgoLined
 Hyper-parameters $T_g, K, \eta_l, \eta_g$\;
 \For{\emph{each group} $g=1,2,\ldots$ \emph{in parallel}}{
    Initialize $\Theta_{g,1}:=\Theta_{T+1}$\;
     \For{\emph{each round} $t=1,2,\ldots, T_g$}{
      Deliver $\Theta_{g,t}$ to clients in group $g$\\
      Sample a subset $\mathcal{I}_{g,t}$ of clients\\
      \For{\emph{each client} $i\in\mathcal{I}_{g,t}$ \emph{in parallel}}{
        Load $\theta_{i,1}^t:=\Theta_{g,t}$\\
        Train $K$ epochs via \texttt{SGD}($\theta_{i,k}^t$, $\eta_l$)\\
        Send $\Delta_i^t:=\Theta_{g,t}-\theta_{i,K+1}^t$\\
      }
      $\Theta_{g, t+1}\leftarrow\texttt{FedAdam}(\Theta_{g,t}, \Delta_{\mathcal{I}_{g,t}}, \eta_g)$\\
     }
     Emit $\Theta_{g,T_g+1}$;
 }
 \caption{Group FL Fine-Tuning.}
 \label{algo2}
\end{algorithm}

\subsection{Local Personalization}
\label{method:3}
Given the group-specific model of $\Theta_{g, T_g+1}$ from the group FL fine-tuning step above, local personalization is then performed using the private training data of each client. Specifically, for any client $i$ in group $g$, we use $\Theta_{g, T_g+1}$ as the seed model and train $K_l$ epochs on local data via \texttt{SGD} with $\eta_{i,l}$ as the learning rate. The resulting model, $\Theta_{i, T_g+1,K_l+1}$, will be adopted for inference. This local personalization step is outlined in Algorithm~\ref{algo3}.

\begin{algorithm}
\SetAlgoLined
 Hyper-parameters $K_l, \eta_{i,l}$; \\
 \For{\emph{each group} $g=1,2,\ldots$ \emph{in parallel}}{
     \For{\emph{each client} $i$ \emph{in group }$g$ \emph{in parallel}}{
        Load $\theta_{i,1}^t:=\Theta_{g,T_g+1}$\\
        Train $K_l$ epochs via \texttt{SGD}($\theta_{i,k}^t$, $\eta_{i,l}$)\\
        Emit $\Theta_{i,T_g+1,K_l+1}:=\theta_{i,K_l+1}^t$;
     }
 }
 \caption{Local Personalization.}
 \label{algo3}
\end{algorithm}

\section{A Bayesian View}
\label{properties}
In this section, we discuss the theoretical insights of group personalized FL from a Bayesian perspective. Consider the following hierarchical model
\begin{align*}
\theta_0 &\sim \pi_0(\cdot) \\
\theta_m \;|\; \theta_0 &\overset{\text{iid}}{\sim} \mathcal{N}(\theta_0, \sigma_0^2),\,\,m=1,\ldots, M \\
\theta_{mn} \;|\; \theta_m &\overset{\text{iid}}{\sim} \mathcal{N}(\theta_m, \sigma_{m}^2),\,\,n=1,\ldots, N_m \\
x_{mn} \;|\; \theta_{mn} &\overset{\text{iid}}{\sim} \mathcal{N}(\theta_{mn}, \sigma_{mn}^2)
\end{align*}
where $\sigma_0^2$, $\sigma_m^2$, $\sigma_{mn}^2$ are fixed constants, $\pi_0(\cdot)$ represents a non-informative flat prior, $\mathcal{N}(\mu,\sigma^2)$ is a Gaussian distribution with mean $\mu$ and variance $\sigma^2$. Here, $x_{mn}$ refers to the data of the $n$th client in the $m$th group, which is distributed according to a client-specific parameter $\theta_{mn}$; each $\theta_{mn}$ in the $m$th group follows a distribution decided by a group-specific parameter $\theta_{m}$; the global parameter $\theta_{0}$ governs the distribution of these $\theta_{m}$'s.

For the sake of simplicity, we assume $\theta_{mn}=\theta_{m}$ in this study and thus different clients in the $m$th group only differs on the variance $\sigma_{mn}^2$. Without the loss of generality, we study the local model of the client parameterized by $\theta_{11}$. The following computes its posterior distributions under different level of knowledge sharing.

\textbf{Without knowledge sharing}. When the target client can only access their own local data, we have the posterior distribution of $\theta_{11}$ written as
\begin{align*}
    \theta_{11} \;|\; x_{11} \sim \mathcal{N}(x_{11}, \sigma_{11}^2)
\end{align*}

\textbf{With group knowledge sharing}. If the target client is able to learn knowledge and data information from other clients in the same group, it can be shown that
\begin{align*}
&\theta_{11} \;|\; x_{11},\ldots,x_{1N_1} \sim \mathcal{N}(\mu_{g1}, \sigma_{g1}^2) \\
&\mu_{g1}: = \frac{\sum_{n=1}^{N_1}\sigma_{1n}^{-2}x_{1n}}{\sum_{n=1}^{N_1}\sigma_{1n}^{-2}}, \,\,\sigma_{g1}^2:=\frac{1}{\sum_{n=1}^{N_1}\sigma_{1n}^{-2}}
\end{align*}
The ratio of posterior variances below measures the reduced statistical uncertainty conditional on other clients' data in the same group
\begin{align*}
    \frac{\sigma_{g1}^2}{\sigma_{11}^2}=\frac{1}{1 + \sum_{n=2}^{N_1}\,(\sigma_{11}/\sigma_{1n})^{2}}< 1
\end{align*}

\textbf{With global knowledge sharing}. In this case, the target client can benefit from all other clients, then we obtain
\begin{align*}
&\theta_{11} \;|\; x_{11},\ldots,x_{MN_M} \sim \mathcal{N}(\mu_{G1}, \sigma_{G1}^2) \\
&\mu_{G1}: = \frac{\sum_{n=1}^{N_1}\sigma_{1n}^{-2}x_{1n}+(\sigma_0^2+\sigma_{G/1}^2)^{-1}\mu_{G/1}}{\sum_{n=1}^{N_1}\sigma_{1n}^{-2}+(\sigma_0^2+\sigma_{G/1}^2)^{-1}} \\
& \sigma_{G1}^2:=\frac{1}{\sum_{n=1}^{N_1}\sigma_{1n}^{-2}+(\sigma_0^2+\sigma_{G/1}^2)^{-1}}
\end{align*}
where
\begin{align*}
&\mu_{G/1}=\frac{\sum_{m=2}^M\sum_{n=1}^{N_m}(\sigma_0^2+\sigma_{mn}^2)^{-1}x_{mn}}{\sum_{m=2}^M\sum_{n=1}^{N_m}(\sigma_0^2+\sigma_{mn}^2)^{-1}} \\
&\sigma_{G/1}^2=\frac{1}{\sum_{m=2}^M\sum_{n=1}^{N_m}(\sigma_0^2+\sigma_{mn}^2)^{-1}}
\end{align*}
The ratio of posterior variances between $\sigma_{G1}^2$ and $\sigma_{g1}^2$ is given by
\begin{align*}
\frac{\sigma_{G1}^2}{\sigma_{g1}^2}=\frac{1}{1 + \frac{(\sigma_0^2+\sigma_{G/1}^2)^{-1}}{ \sum_{n=1}^{N_1}\sigma_{1n}^{-2}}} < 1    
\end{align*}
which quantifies the additional reduced uncertainty conditional on clients' data information over all groups.

To summarize, comparing the posterior variances among the settings of (1) without knowledge sharing, (2) with group knowledge sharing, and (3) with global knowledge sharing, we can see that higher level of knowledge sharing results in lower uncertainty on unknown parameter $\theta_{11}$.

\section{Experiments}
\label{experiments}
\subsection{Datasets}
We first evaluate the proposed method on the transcripts of in-house video dataset, which is sampled from public social media videos and de-identified before transcription; both transcribers and researchers do not have access to any user-identifiable information. This data can be partitioned into several categories based on different topics or genres of videos uploaded by the owners. Table~\ref{tab:video} shows the training and evaluation splits as well as the summary statistics on numbers of owners (i.e.~uploaders), videos, and words in each video category.

\begin{table}[ht]
  \caption{Summary statistics of the video dataset.}
  \centering
  \resizebox{\columnwidth}{!}{%
  \begin{threeparttable}
  \begin{tabular}{lc|rr|rr}
    \toprule
    &
    & \multicolumn{2}{|c}{\emph{Training}} 
    & \multicolumn{2}{|c}{\emph{Evaluation}}\\
    \cmidrule(r){3-4}
    \cmidrule(r){5-6}
    \emph{Category}& \emph{\#owners}& \emph{\#videos} & \emph{\#words} & \emph{\#videos} & \emph{\#words} \\
    \midrule
    \texttt{general} & 9178 & 16769 & 1074K & 10764 & 597K \\
    \texttt{ads} & 5905 & 22764 & 3459K & 10658 & 1647K \\
    \texttt{podcast} & 103 & 825 & 229K & 325 & 94K \\
    \texttt{football} & 28 & 59 & 20K & 34 & 12K \\
    \texttt{news} & 12 & 302 & 93K & 105 & 34K \\
    \texttt{gaming} & 11 & 112 & 6K & 45 & 4K \\
    \texttt{basketball} & 6 & 98 & 6K & 36 & 2K \\
    \bottomrule
  \end{tabular}
  \end{threeparttable}
  }
  \label{tab:video}
  \vspace{-0.3cm}
\end{table}

We then experiment with Wikitext-103 data \cite{merity2016pointer}. For each wiki page, we partition the text corpus into sentences; 75\% of sentences are allocated for training, and the remaining 25\% are for evaluation. Particularly, we select 5 topic categories as a subset of all wiki pages, according to the titles of their pages. Table~\ref{tab:wiki} displays the summary statistics on numbers of wiki pages, sentences, and words in each topic category of wiki pages, for both training and evaluation splits.

\begin{table}[ht]
  \caption{Summary statistics of the wiki dataset.}
  \centering
  \resizebox{\columnwidth}{!}{%
  \begin{threeparttable}
  \begin{tabular}{lc|rr|rr}
    \toprule
    &
    & \multicolumn{2}{|c}{\emph{Training}} 
    & \multicolumn{2}{|c}{\emph{Evaluation}}\\
    \cmidrule(r){3-4}
    \cmidrule(r){5-6}
    \emph{Category}& \;\,\emph{\#pages}\; & \;\;\emph{\#sents} & \emph{\#words} & \;\;\emph{\#sents} & \emph{\#words} \\
    \midrule
    \texttt{all} & 28726 & 1263K & 65606K & 421K & 21095K\\
    \midrule
    \texttt{battle} & 409 & 22641 & 1164K & 7557 & 377K \\
    \texttt{film} & 336 & 21408 & 1125K & 7137 & 367K \\
    \texttt{video game} & 137 & 5449 & 282K & 1811 & 90K \\
    \texttt{music} & 53 & 2427 & 124K & 812 & 40K \\
    \texttt{disease} & 10 & 845 & 41K & 281 & 14K \\
    \bottomrule
  \end{tabular}
  \end{threeparttable}
  }
  \label{tab:wiki}
  \vspace{-0.3cm}
\end{table}

\subsection{Setups}
We conduct the language model (LM) task in our experiments. The LM is LSTM based with character embedding \cite{kim2016character} dimension 100, and 2 layers of 512 hidden units. The vocabulary size is around 33K (at word level).

To simulate the FL environment for the video dataset, each video uploader is treated as a client and their videos are considered as training or evaluation examples. Each owner only uploads one category of videos and thus the clients can be clustered into different groups according to their categories of videos. For the wiki dataset, each wiki page is treated as a client and the corresponding sentences are considered as training or evaluation examples. The clients are grouped based on the topics of their wiki pages.

Regarding the hyper-parameters of global and group FL training, we set the number of selected users per round $|\mathcal{I}_t|=|\mathcal{I}_{g,t}|=100$; learning rate $\eta_G=\eta_g=0.001$ in the global \texttt{FedAdam} optimizer and $\eta_l=1.0$ for the client \texttt{SGD} optimizer. Locally, we train $K=1$ epoch with batch size 8 for any selected client per FL round. We use $T=20$ epochs for global FL training and $T_g=10$ for group FL fine-tuning. For personalization, we set $K_l=5$ and learning rate $\eta_{i,l}$ varies at $0.001$, $0.01$, $0.1$, and $1.0$.

All experiments are performed using the open-source FL simulation framework ``FLSim'' \cite{flsim2022}.

\subsection{Methods in Comparison}
In our experiments, we consider the following methods
\begin{itemize}
\item \textsf{FL}. This is the vanilla version of FL training \cite{konevcny2016federated, konevcny2016federated2, mcmahan2017communication} on LMs. Specifically, for the video dataset, LM is trained via FL using the train split on 7 categories of video transcripts; for the wiki dataset, LM is trained on the train split of all Wikitext-103 corpus (including these 5 categories of topics in Table~\ref{tab:wiki});
\item \textsf{PerFL}. This is the fine-tuning based personalization with the global FL trained model as the seed model \cite{tan2022towards}; 
\item \textsf{GroupFL}. This is the group FL fine-tuning approach which is essentially described in Algorithms~\ref{algo1}-\ref{algo2}. It is worth noting that it can be considered as a comparable but stronger baseline than the cluster FL method \cite{ghosh2020efficient};
\item \textsf{GroupPerFL}. This is our proposed method (Algorithms~\ref{algo1}-\ref{algo3}).
\end{itemize}

\subsection{Results on Video Dataset}
% Briefly mention we evaluate with WER and perplexity as a first sentence
Table~\ref{tab:2} displays the perplexity results, aggregated at sentence level, on the evaluation split of video dataset. From these results, we can observe that \textsf{GroupPerFL} clearly outperforms other methods, and particularly obtains 5\%-20\% perplexity reduction on most categories of videos in comparison with \textsf{PerFL}. Notice that leveraging group-level information is more beneficial when a majority of users have smaller numbers of training examples. For instance, the \texttt{general} category has lower ratio of \emph{\#videos} to \emph{\#owners} than the \texttt{ads} category, and achieves larger improvement comparing \textsf{GroupFL} with \textsf{FL}. This is expected since group-level knowledge sharing is particularly helpful when the training records per client are insufficient.

%comparing the methods of global FL training (\textsf{FL}), two-step fine-tuning based personalized FL (\textsf{PerFL}), group FL fine-tuning (\textsf{GroupFL}), and proposed three-step group personalized FL (\textsf{GroupPerFL}).

\begin{table}[ht]
  \caption{Perplexity results on the video evaluation dataset.}
  \centering
  \resizebox{\columnwidth}{!}{%
  \begin{threeparttable}
  \begin{tabular}{l|r|r|r|r}
    \toprule
    \emph{Category}& \textsf{$\,\qquad$ FL} & \textsf{$\quad$ PerFL} & \textsf{GroupFL} & \textsf{GroupPerFL} \\
    \midrule
    \texttt{general} & 211.2 & 190.1 & 178.3 & \textbf{168.1} \\
    \texttt{ads} & 169.3 & 124.7 & 158.2 & \textbf{117.6} \\
    \texttt{podcast} & 162.4 & \textbf{110.4} & 140.4 & 110.6 \\
    \texttt{football} & 209.5 & 207.5 & 208.2 & \textbf{207.2} \\
    \texttt{news} & 253.6 & 224.1 & 226.0 & \textbf{218.7} \\
    \texttt{gaming} & 233.9 & 185.9 & 199.3 & \textbf{168.7} \\
    \texttt{basketball} & 470.7 & 290.7 & 272.8 & \textbf{228.6} \\
    \bottomrule
  \end{tabular}
  \end{threeparttable}
  }
  \label{tab:2}
\end{table}

\begin{figure}[ht]
  \centering
  \includegraphics[width=\linewidth]{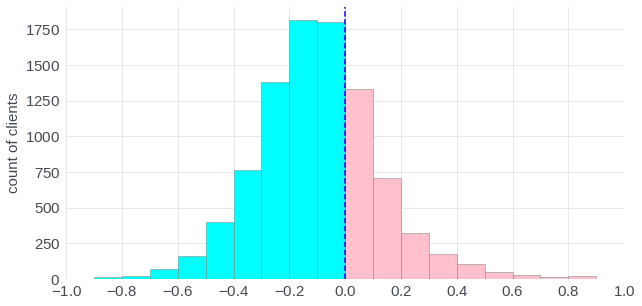}
  \caption{Histogram on relative perplexity change at client level, comparing \textsf{GroupPerFL} with \textsf{PerFL} on the \texttt{general} category of video evaluation dataset.}
  \label{fig:hist}
\end{figure}

For the \texttt{general} category of video evaluation dataset, Figure~\ref{fig:hist} shows the histogram on client-level relative perplexity change comparing \textsf{GroupPerFL} against \textsf{PerFL}, where we can see approximately 70\% of clients have reduced perplexity.

For better measuring the quality-cost trade-offs, Table~\ref{tab:3} displays the total communication cost and computation cost of each client in comparison of these methods.

\begin{table}[!ht]
  \caption{Results on communication and computation costs at client level for \texttt{general} video evaluation data; unit of communication is model size, while unit of computation is the cost per each epoch of local training.}
  \centering
  \resizebox{\columnwidth}{!}{%
  \begin{threeparttable}
  \begin{tabular}{l|r|r|r|r}
    \toprule
    \emph{Metric}& \textsf{$\quad\qquad$ FL} & \textsf{$\,\,\quad$ PerFL} & \textsf{$\,$ GroupFL} & \textsf{GroupPerFL} \\
    \midrule
    \emph{Communication} & 3060 & 3060 & 3990 & 3990 \\
    \emph{Computation} & 20 & 25 & 30 & 35 \\
    \bottomrule
  \end{tabular}
  \end{threeparttable}
  }
  \label{tab:3}
  \vspace{-0.3cm}
\end{table}

Since the video dataset comes with audios, we also evaluate the speech recognition performance with the LMs being used as second-pass rescorers on the generated 20-best hypotheses. The ASR model is an RNN-T model with the Emformer encoder \cite{emformer2021streaming}, LSTM predictor, and a joiner. It has around 80 million parameters and is trained from scratch using the training utterances of entire video data. The LM interpolation weight is set to 0.10 across all methods. Table~\ref{tab:wer} shows the word error rate (WER) results on the \texttt{general} category of video evaluation dataset, where we can see \textsf{GroupPerFL} achieves the best speech recognition quality, 2.8\% relative WER improvement compared with the baseline ASR without LM rescoring.

\begin{table}[ht]
  \caption{WER results on the \texttt{general} category of video evaluation dataset, using LMs as second-pass rescorers.}
  \centering
  \resizebox{\columnwidth}{!}{%
  \begin{threeparttable}
  \begin{tabular}{c|r|r|r|r}
    \toprule
    \emph{Baseline ASR}& \multicolumn{4}{|c}{\emph{ASR w/ 2nd-Pass LM Rescoring}} \\
    \cmidrule(r){2-5}    
    \emph{(w/o LM)}& \textsf{$\,\qquad$ FL} & \textsf{$\quad$ PerFL} & \textsf{GroupFL} & \textsf{GroupPerFL} \\
    \midrule
    26.83 & 26.56 & 26.16 & 26.32 & \textbf{26.09} \\
    \bottomrule
  \end{tabular}
  \end{threeparttable}
  }
  \label{tab:wer}
  \vspace{-0.3cm}
\end{table}

\subsection{Results on Wiki Dataset}
Table~\ref{tab:wiki_eval} shows the perplexity results on the evaluation split of wiki dataset regarding the 5 categories of topics. Again, we can notice that utilizing group information is useful from the comparison between \textsf{GroupFL} and vanilla \textsf{FL}, and also the \textsf{GroupPerFL} method achieves 2\%-12\% perplexity improvement on all the categories over \textsf{PerFL}. Here, such gains are relatively smaller compared with the ones observed in the video dataset since each client (i.e. wiki page) has more training examples so that performing personalization using their own data could be already good enough, while utilizing group-level knowledge is still helpful but less beneficial than the scenarios in the video dataset. This can also be verified by the observation that \textsf{PerFL} outperforms \textsf{GroupFL} for all the topics.  

\begin{table}[ht]
  \caption{Perplexity results on the wiki evaluation dataset.}
  \centering
  \resizebox{\columnwidth}{!}{%
  \begin{threeparttable}
  \begin{tabular}{l|r|r|r|r}
    \toprule
    \emph{Category}& \textsf{$\,\qquad$ FL} & \textsf{$\quad$ PerFL} & \textsf{GroupFL} & \textsf{GroupPerFL} \\
    \midrule
    \texttt{battle} & 89.8 & 67.4 & 79.4 & \textbf{65.6} \\
    \texttt{film} & 124.1 & 101.7 & 108.0 & \textbf{98.5} \\
    \texttt{video game} & 134.4 & 109.6 & 111.8 & \textbf{103.2} \\
    \texttt{music} & 118.1 & 74.2 & 92.1 & \textbf{65.6} \\
    \texttt{disease} & 140.1 & 97.2 & 106.5 & \textbf{92.2} \\
    \bottomrule
  \end{tabular}
  \end{threeparttable}
  }
  \label{tab:wiki_eval}
  \vspace{-0.3cm}
\end{table}

\section{Conclusion}
\label{conclusion}
We present group personalized FL for integrating global aggregation, group-level knowledge sharing, and local personalization. The proposed approach could be well interpreted from a Bayesian hierarchical modeling perspective. We demonstrate our method is effective in achieving improved personalization results through experiments on two real-world datasets for language modeling task.

\bibliographystyle{IEEEbib}
\bibliography{refs}

\end{document}